\documentclass{article}
\usepackage{spconf,amsmath,epsfig, amsfonts}

\usepackage{algorithm}
\usepackage{algorithmic}

\let\OLDthebibliography\thebibliography
\renewcommand\thebibliography[1]{
  \OLDthebibliography{#1}
  \setlength{\parskip}{0pt}
  \setlength{\itemsep}{0pt plus 0.3ex}
}

\begin{document}\sloppy

\def\x{{\mathbf x}}
\def\L{{\cal L}}

\title{Client-supervised Federated Learning: Towards One-model-for-all Personalization }
%
\name{Peng Yan\footnote{yanpeng9008@hotmail.com}, Guodong Long\footnote{guodong.long@uts.edu.au}}
\address{}

\maketitle

\begin{abstract}
Personalized Federated Learning (PerFL) is a new machine learning paradigm that delivers personalized models for diverse clients under federated learning settings. Most PerFL methods require extra learning processes on a client to adapt a globally shared model to the client-specific personalized model using its own local data. However, the model adaptation process in PerFL is still an open challenge in the stage of model deployment and test time. This work tackles the challenge by proposing a novel federated learning framework to learn only one robust global model to achieve competitive performance to those personalized models on unseen/test clients in the FL system. Specifically, we design a new \textbf{Client-Supervised Federated Learning (FedCS)} to unravel clients' bias on instances' latent representations so that the global model can learn both client-specific and client-agnostic knowledge. Experimental study shows that the FedCS can learn a robust FL global model for the changing data distributions of unseen/test clients. The FedCS's global model can be directly deployed to the test clients while achieving comparable performance to other personalized FL methods that require model adaptation.
\end{abstract}
\section{Introduction }
\label{Introduction}
Modern machine learning relies on massive data to train models like deep neural networks (DNNs), but collecting data is becoming more sensitive with increasing attention to privacy protection. Then, federated learning (FL)~\cite{mcmahan2017communication} emerged and became a popular learning paradigm in recent years. The FL aims to coordinate a set of clients (i.e., devices) to train a global model while preserving their data locally and privately. Further, personalized FL (PerFL)~\cite{deng2020adaptive,mansour2020three,wu2022motley} balances cross-client training and personalization. Clients will collaborate as in vanilla FL and leverage client-specific properties to learn personalized models demonstrating superior performance on non-IID data.

Existing personalized federated learning research usually aims to learn many client-specific personalized models to tackle the non-IID data in federated learning systems. Specifically, a global model will be learned to grasp shared knowledge, and then many personalized models will be learned or fine-tuned on the client by leveraging the global model and local dataset. Although this client-specific personalised model strategy has the potential to catch non-IID in fine-grained, the on-device learning and fine-tuning process is usually difficult to control in practice, especially in the stage of model deployment and test time.

This paper aims to rethink the personalized federated learning problem by proposing a one-model-for-all strategy to embody personalization in federated settings. One motivation is that model personalization in most PerFL methods relies on the bias of instances on the same client, and little supervised information describing the client is introduced. Then, the global model trained on the same data will attain the same performance if it can recognize the bias of a client. Inspired by this, we propose to use a one-model-for-all strategy to learn a unified model that can be shared across clients in the FL system. Moreover, the client-specific information will be encoded in a unified representation space and then be fed into a decision module along with the client-agnostic knowledge to make the final prediction.

\begin{figure}[t]
  \centerline{\epsfig{figure=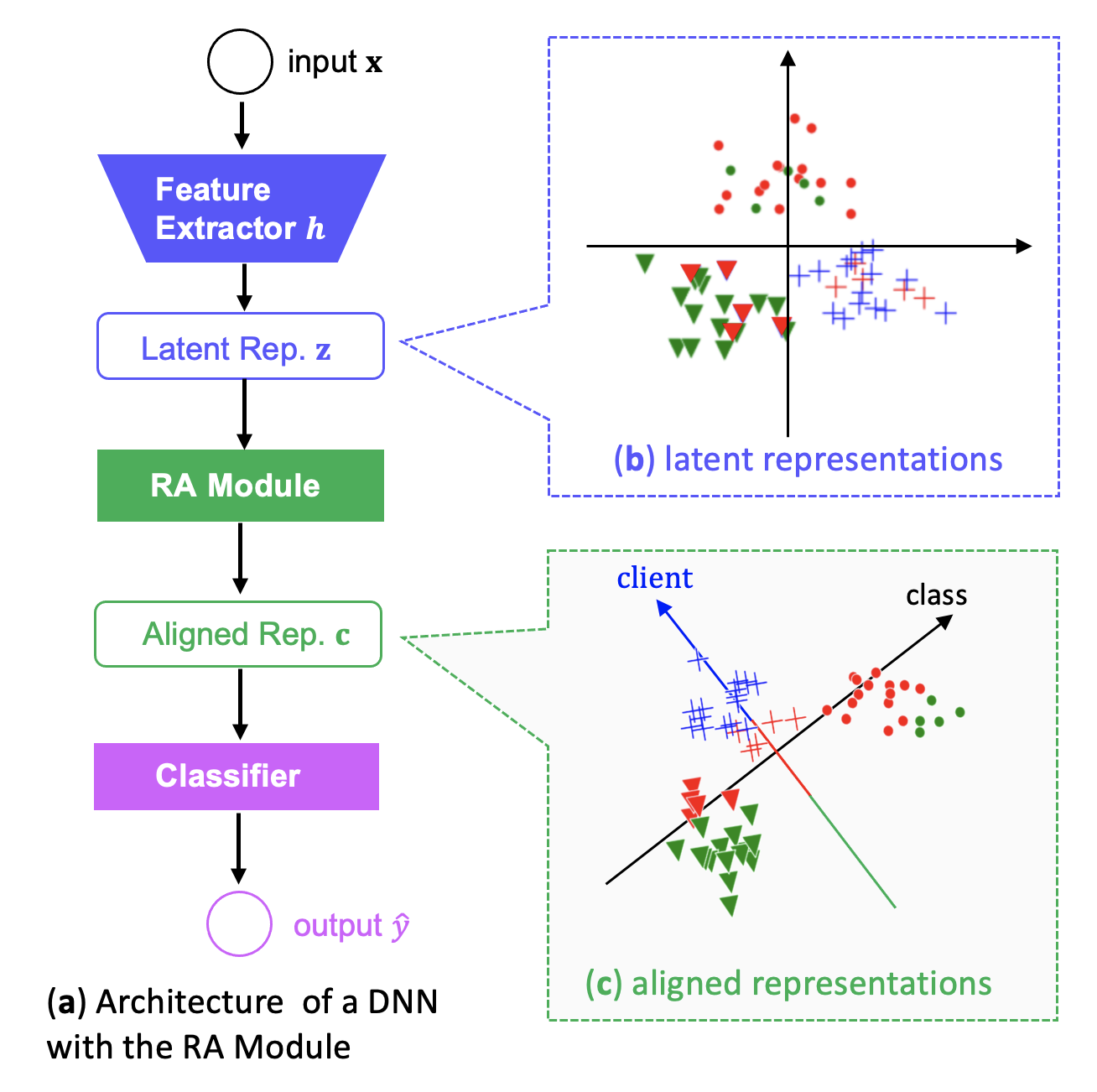,width=8.5cm}}
%
\caption{Illustration to the FedCS. (b) and (c) describe distributions of instances' latent representations before and after the Representation Alignment module (RA Module). Types of markers denote classes, and colors indicate clients they are observed. The vanilla feature extractor is unable to recognize clients' bias. RA module in FedCS will align the hidden space so that latent representations can indicate biases of clients.}
\label{fig:architecture}
\end{figure}

Based on the thought above, this work proposes a novel \textbf{Client-Supervised Federated Learning (FedCS)} that is to learn a unified global model with the below functions, including 1) to learn a unified representation space that can encode the bias of a client, 2) to share client-agnostic knowledge as vanilla FL methods, and 3) to make personalized predictions by leveraging both pieces of information. Moreover, the \textbf{Representation Alignment (RA)} mechanism in FedCS could become a plug-in component to be integrated with any federated learning methods. It enables a vanilla FL model to output personalized results without on-device fine-tuning steps. An illustration of the FedCS is in~Fig.\ref{fig:architecture}.


To learn the above objectives in a federated optimization framework, we designed a novel \textbf{Client-Supervised Federated Optimization Framework} to align the objective representation space while being consistent with the federated optimization framework. It exploits the bias that instances on the same client are influenced by identical client properties and formulates the representation aligning task into an optimization problem that clients can solve collaboratively. 

Through qualitative and quantitative experiments, we illustrate how FedCS is integrated into a black-box model to achieve compared performance of other personalized federated learning methods. 
Moreover, we demonstrate that FedCS is compatible with most FL models. Vanilla models with an RA module can achieve competitive performance compared to those ad hoc PerFL models while not need extra fine-tuning steps or personal parameters.

The main contributions are summarized as follows:
\begin{itemize}
\item We propose a novel one-model-for-all personalized FL framework that won't require an extra fine-tuning process at the stage of model deployment and test time. The personalization in the FL system is carried on representations indicating client bias rather than models.

\item We designed a novel \textbf{representation alignment} mechanism to project instances' representation into a space indicating clients' biases. The following decision layers in the neural architecture can automatically learn to make personalized predictions by leveraging the client's bias.

\item A \textbf{client-supervised optimization framework} is designed to fit the proposed framework. It formulates the representation-aligning problem into a unified optimization framework that clients can solve collaboratively under FL settings.

\item Contrast with baseline methods shows that, by integrating FedCS into vanilla FL models, they can achieve competitive personalization performance without requiring extra fine-tuning steps or personal parameters.
\end{itemize}

\section{Related Work}
\label{Personalized Federated Learning}
\textbf{Federated Learning} is an emerging machine learning paradigm where many clients collaborate to train a global model by sharing and aggregating model parameters rather than user data~\cite{mcmahan2017communication}. 
Further work~\cite{deng2020adaptive,mansour2020three,wu2022motley} shows that the global model performs better after tuning toward client-specific properties and proposes \textbf{Personalized Federated Learning}.

A popular way to personalize is to fine-tune the global model on a client's local data~\cite{mcmahan2017communication,cheng2021fine,collins2022fedavg}; meta-learning strategies will help improve the tuning process~\cite{jiang2019improving,fallah2020personalized}. Meanwhile, \cite{li2021ditto} trains a local model for each client while constraining the distributed training procedures with a shared regularizer. 
Recent work~\cite{pillutla2022federated} studies partial personalization, splitting an FL model into global and local parts. Personalization is fulfilled by individually training the local part of a model on clients. \cite{li2021fedbn} is a simple but efficient case of these methods. It trains a model through the vanilla FedAvg~\cite{mcmahan2017communication} except for preserving batch-normalization modules locally.~\cite{collins2021exploiting} introduces a method to learn a shared data representation across clients and unique classification heads for each client. \cite{luo2022disentangled} utilizes a global and a local encoder to learn different representations for cross-client collaboration and personalization. \cite{ye2023upfl} and \cite{sun2022feature} propose to learn unique representations for client properties to improve personalization on new clients. However, these methods require extra training data to capture those clients' properties when deployed on them, which is only sometimes feasible.

\section{Problem Formulation}
\label{Problem Formulation}
\textbf{Federated Learning (FL)} assumes $K$ clients participate in a learning process with ${\{}\mathbf{x}_1^{(i)}, \mathbf{x}_2^{(i)},...,\mathbf{x}_{N_i}^{(i)} \}\in \mathcal{X}^{(i)}$ denoting instances on the $i$-th client, and $\{y_1^{(i)},y_2^{(i)},...,y_{N_i}^{(i)}\}\in \mathcal{Y}^{(i)}$ are their labels. The FL task is to find the optimal parameters $\omega^{*}$ for a global model $f(\mathbf{x};\omega)$ by minimising the total loss of all clients as the following optimization problem:


\begin{equation}
    \label{def-fl}
    \begin{aligned}
    \omega^{*}=\arg\min_{\omega}\sum_{i=1}^{K}\alpha_i \mathcal{L}_i(\omega)
    \end{aligned}
\end{equation}

where $\mathcal{L}_i(\omega)=(1/N_i)\sum_{j=1}^{N_i}l(f(\mathbf{x}_j^{(i)};\omega),y_{j}^{(i)})$ is the supervised loss on the $i$-th client, and $\alpha_i$ is its weight. In particular, in the standard FL framework~\cite{mcmahan2017communication}, $\alpha_i$ is the fraction of the size of the client's training data, i.e., $\alpha_i= N_i/\sum_{i'=1}^{K}N_{i'}$. Each client will update $\omega$ privately by minimizing $\mathcal{L}_i(\omega)$, and a server will orchestrate the distributed training process by collecting locally updated $\omega$ and synchronizing the averaged one.

\textbf{Personalized Federated Learning (PerFL)} leverages cross-client collaboration but learns a personalized model $f(\mathbf{x};\omega,\mu_i)$ for each client, where $\omega$ denotes parameters shared, and $\mu_i$ denotes parameters for the $i$-th client. The learning task can be formulated into a unified optimization problem~\cite{pillutla2022federated}:
\begin{equation}
    \label{def-perfl}
    \begin{aligned}
    \omega^{*},\{\mu_i^{*}\}_{i=1}^{K}=\arg\min_{\omega,\{\mu_i\}_{i=1}^{K}}\sum_{i=1}^{K}\alpha_i\mathcal{L}_i(\omega, \mu_i)
    \end{aligned}
\end{equation}
There are different types of personalization according to how to define $\omega$ and $\mu_i$. 1) \textbf{full personalization}: a local model is fully parameterized by $\mu_i$, and the global model guides the local training process, e.g., regularizing $\mu_i$ by minimizing $\|\omega-\mu_i\|_2$~\cite{li2021ditto} or initializing local parameters with $\mu_i= \omega-\nabla\mathcal{L}_i$~\cite{fallah2020personalized}; 2) \textbf{partial personalization}: $\omega$ and $\mu_i$ constitute the global and personal parts of a local model, where $\omega$ is shared through the fundamental FL method in \ref{def-fl}, and $\mu_i$ is trained individually on each client. e.g., $\omega$ could be parameters of a shared backbone model, and $\mu_i$ is a classification head for the $i$-th client~\cite{collins2021exploiting}. 

\section{Methodology}
\label{Methodology}
Looking inside the latent space of a DNN $f(\mathbf{x};\omega)=g(h(\mathbf{x};\omega_h);\omega_g)$, it consists of two parts: $h(\mathbf{x};\omega_h)$ is a feature extractor learning the latent representation $\mathbf{z}\in\mathbb{R}^{d}$, and $g(\mathbf{z};\omega_g)$ is a classification head making predictions based on $\mathbf{z}$. Our goal is to align the latent representation space for $\mathbf{z}$, such that 1) it is able to embed client-specific information, i.e., instances from similar clients will have similar values and vary significantly otherwise; 2) it is able to share client-agnostic knowledge as vanilla FL methods, and 3) downstream modules can make personalized predictions by leveraging both information.


Formally, FedCS looks for a projection $\mathbf{c}=\mathbf{P}^{T}\mathbf{z}$, where $\mathbf{P}_{d\times r}=[\mathbf{p}_1,\mathbf{p}_2,...,\mathbf{p}_r]$ is the orthonormal basis of the objective representation space. It searches for the optimal directions $\mathbf{P}^{*}$ according to an inductive bias that representation $\mathbf{c}$ on the same client shall be similar, and those from different clients are on the contrary.

\subsection{Representation Alignment}
Specifically, let $\bar{\mathbf{c}}^{(i)}$ denote the mean of representations on the $i$-th client, and $\bar{\mathbf{c}}$ be the global mean among clients.
\begin{equation}
    \begin{aligned}
    \Sigma_W=\frac{1}{\sum_{i=1}^{K}N_{i}}\sum_{i=1}^{K}\sum_{j=1}^{N_i}(\mathbf{c}_j^{(i)}-\bar{\mathbf{c}}^{(i)})(\mathbf{c}_j^{(i)}-\bar{\mathbf{c}}^{(i)})^{T}
    \end{aligned}
    \label{def-within-client-scatter-matrix}
\end{equation}
Eq.\ref{def-within-client-scatter-matrix} is the within-client scatter matrix that measures the scatter of latent representations within each client and
\begin{equation}
    \begin{aligned}
    \Sigma_B=\frac{1}{\sum_{i=1}^{K}N_{i}}\sum_{i=1}^{K}N_i(\bar{\mathbf{c}}^{(i)}-\bar{\mathbf{c}})(\bar{\mathbf{c}}^{(i)}-\bar{\mathbf{c}})^{T}
    \end{aligned}
    \label{def-between-client-scatter-matrix}
\end{equation}
Eq.\ref{def-between-client-scatter-matrix} is the between-client scatter matrix that measures the scatters of the mean across clients. To find the $\mathbf{P}^{*}$ is to find the directions that minimize $\Sigma_W$ and maximize $\Sigma_B$. For example, it can be formulated as the Linear Discriminate Analysis (LDA) problem below
\begin{equation}
    \mathbf{P}^{*}=\arg\max_{\mathbf{P}}J(\mathbf{P})=\arg\max_{\mathbf{P}}\text{Tr}(\Sigma_W^{-1}\Sigma_B)
    \label{def-rayleigh-quotient}
\end{equation}
where $\text{Tr}(\cdot)$ denotes the trace of the matrix.

Then, bring Eq.\ref{def-rayleigh-quotient} into the FL framework, the overall learning task is formulated as a bi-level optimization problem
\begin{equation}
    \begin{aligned}
    \omega_h^{*},\omega_g^{*} ~&=\arg\min_{\omega_h,\omega_g}\sum_{i=1}^{K}\alpha_i\mathcal{L}_i(\omega_h,\omega_g)\\
    s.t.~&\mathbf{P}^{*}=\arg\max_{\mathbf{P}}J(\mathbf{P})\\
    \label{def-overall-loss}
    \end{aligned}
\end{equation}
where 
\begin{equation}
    \begin{aligned}
    \mathcal{L}_i=\sum_{j=1}^{N_{i}}l(g(\mathbf{P}^{T}h(\mathbf{x}_j^{(i)};\omega_h);\omega_g),y_j^{(i)})\\
    \label{def-overall-loss-local}
    \end{aligned}
\end{equation}
In the next section, we introduce a client-supervised method to optimize the Eq.\ref{def-overall-loss} under the FL setting.

\subsection{Client Supervised Optimization}
Theoretically, the optima of Eq.\ref{def-rayleigh-quotient} are eigenvectors of $\Sigma_{W}^{-1}\Sigma_{B}$ associated with the $r$ largest eigenvalues~\cite{duda2006pattern}. Then, the classification and the alignment tasks in Eq.\ref{def-overall-loss} can be optimized alternatively under the conventional FL framework~\cite{li2021ditto, pillutla2022federated}. However, decomposing $\Sigma_{W}^{-1}\Sigma_{B}$ is computationally expensive, and involves collecting the local mean $\bar{\mathbf{c}}^{(i)}$ which is privacy sensitive. To this end, we propose a client-supervised method that decomposes the learning task in Eq.\ref{def-rayleigh-quotient} into sub-tasks so that clients can optimize it collaboratively. 

Concretely, previous works~\cite{ghassabeh2015fast} show that maximizing Eq.\ref{def-rayleigh-quotient} is equivalent to maximizing $\Sigma_W^{-1/2}\Phi\Sigma_W^{-1/2}$, where $\Phi$ is an approximation to the eigen system of the global correlation matrix $\Sigma_W+\Sigma_B$, and both $\Sigma_W^{-1/2}$ and $\Phi$ can be updated incrementally through the following equations
\begin{equation}
\begin{aligned}
    \Sigma_W^{-1/2} = \Sigma_W^{-1/2} + \eta * (I - \Sigma_W^{-1/2}\Sigma_W\Sigma_W^{-1/2})
    \label{eq-sigma}
\end{aligned}
\end{equation}
and 
\begin{equation}
\begin{aligned}
    \Phi = \Phi + \lambda * (\mathbf{\bar{u}}\mathbf{\bar{u}}^{T}\Phi - \Phi\tau(\Phi\mathbf{\bar{u}}\mathbf{\bar{u}}^{T}\Phi)
    \label{eq-phi}
\end{aligned}
\end{equation}
where $\mathbf{u} = \Sigma_W^{-1/2} \mathbf{\bar{c}}$, and $\tau(\cdot)$ is an operator that sets all the elements below the main diagonal of the matrix to zero. In this process, $\Sigma_W$ summarizes the correlation of instances on each client and hence will work as supervised information to encode the bias of a client into $\mathbf{u}$. Details of the derivations of Eq.\ref{eq-sigma} and Eq.\ref{eq-phi} are discussed in Appendix. Then, the representation alignment process is described in Algorithm.\ref{alg:representation alginment}, and the overall FL process with FedCS is described in Algorithm.\ref{alg:client supervised optimization}.
\begin{algorithm}[tb]
   \caption{Representation Alignment}
   \label{alg:representation alginment}
\begin{algorithmic}
   \STATE {\bfseries Input:} a batch of latent representations $\mathbf{z}$, global mean $\mathbf{z}_{g}$, client's local correlation $\Sigma_W^{-1/2}$ and $\Phi$
   \STATE {\bfseries begin:}
   \STATE 1. if $\Sigma_W^{-1/2}$ and $\Phi$ are empty, initialize $\Sigma_W^{-1/2}$ and $\Phi$
   \STATE 2. calculate local mean: $\mathbf{z}_{l} = mean(\mathbf{z})$
   \STATE 3. update global mean: $\mathbf{z}_{g} = \mathbf{z}_{g} + \frac{1}{|\mathbf{z}|}(\mathbf{z}-\mathbf{z}_{g})$
   \STATE 4. calculate local correlations: $\Sigma_{W}=(\mathbf{z}-\mathbf{z}_{l})(\mathbf{z}-\mathbf{z}_{l})^{T}$
   \STATE 5. update $\Sigma_W^{-1/2}$ by Eq.\ref{eq-sigma}
   \STATE 6. update $\Phi$ by Eq.\ref{eq-phi}
   \STATE 7. $\mathbf{P}=\Sigma_W^{-1/2}\Phi$
   \STATE 8. return $\mathbf{P}$
   \STATE{\bfseries end}
\end{algorithmic}
\end{algorithm}


\begin{algorithm}[tb]
   \caption{Client-Supervised Federate Learning}
   \label{alg:client supervised optimization}
\begin{algorithmic}
   \STATE {\bfseries Input:} training data $(\mathcal{X}^{(i)},\mathcal{Y}^{(i)})$ distributed on $K$ clients, $t$ the round to update $P$.
   \STATE {\bfseries begin:}
   \STATE 1. Initialize the FL model $f(\mathbf{x};\omega)=g(h(\mathbf{x};\omega_h);\omega_g)$.
   \STATE 2. Select a set of clients $\mathbb{S}$
   \STATE {\bfseries for} the $i$-th client in $\mathbb{S}$ {\bfseries parallel do}
   \STATE ~ update $\omega_h$ and $\omega_g$ locally
   \STATE ~ if round\%$t$ == 0:
   \STATE ~ update $\mathbf{P}$ locally by Algorithm.\ref{alg:client supervised optimization}
   \STATE {\bfseries end for}
   \STATE 3. Aggregate local updates of $\omega_h$, $\omega_g$ and $\mathbf{P}$ by averaging
   \STATE{\bfseries end}
\end{algorithmic}
\end{algorithm}

\section{Experiments}
\label{Experiments}
In this section, we demonstrate the advantages of FedCS in learning from clients with non-i.i.d. data. The FedCS can learn a robust FL global model for the changing data distributions of unseen/test clients. The FedCS's global model can be directly deployed to the test clients while achieving comparable performance to other personalized FL methods that require model adaptation.

\subsection{Client Settings}
We simulate FL environments by allocating instances from benchmark datasets to 50 clients, and two types of heterogeneity are applied (Details of client settings are introduced in the Appendix).
\begin{itemize}
\item \textbf{Label-shift}: We experiment, respectively, on the MNIST and the CIFAR-10 datasets. We allocate instances of each label individually according to a posterior of the Dirichlet distribution\cite{hsu2019measuring}, which divides clients into ten groups with different label distributions. Eight groups of clients will participate in the collaborative training process, and the rest will be held for test. 


\item \textbf{Feature-shift}: We experiment on the Digit-5 dataset to evaluate FedCS's performance on feature-shift data. The Digit-5 consists of digits from five domains (MNIST, MNIST-M, SVHN, USPS and Synth Digits). We assign instances of each domain to nine clients, where eight clients will train the global model and one for the test. In addition, we randomly draw instances from all domains to compose five mixed datasets for the test.
\end{itemize}

\subsection{Models and Hyperparameters}
We apply convolution neural networks (CNN) as fundamental models and integrate our proposed RA module into fully connected layers (FCs) to align their hidden layers (see Appendix for details of model architectures). By default, in each communication round, we sample ten clients to update the global model and evaluate the global model's performance on all clients. One epoch of fine-tuning steps will be applied for benchmark methods where the global model must be adapted to a client's local data before testing. The learning rate of a client's local training step is initialized as 0.005, and it will decay at the rate of 0.8 every 50 communication rounds. The RA module will be updated every five communication rounds by the sampled ten clients, and the learning rate is fixed at 0.001. (see Appendix for details of other hyperparameters).



\begin{figure*}[ht]
\begin{center}
\centerline{\includegraphics[width=2.0\columnwidth,]{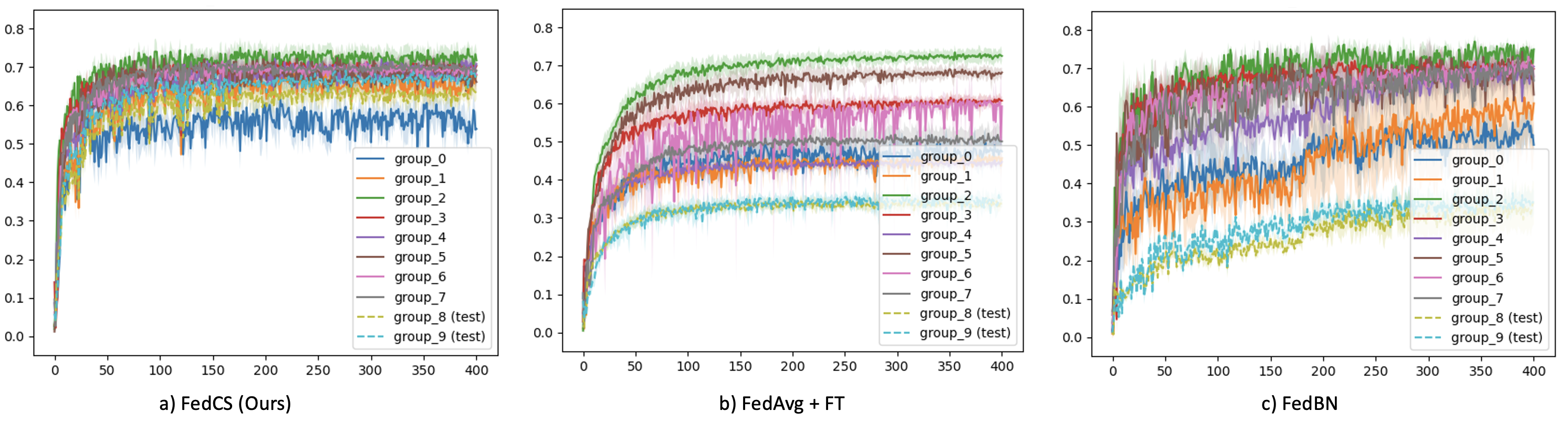}}
\caption{Part of experiments on CIFAR-10. The horizontal axis denotes communication rounds, and the vertical axis denotes F1 Scores. We demonstrate the averaged weighted F1-scores within each client group and marked them with different colours. We can find the performance on different distributions (client groups) vary significantly through FedAvg+FT and FedBN. FedCS has the most robust performance even on unseen clients (group 8 and 9)} 
\label{fig-grouped-f1-score-cifar10}
\end{center}
\end{figure*}

\subsection{Performance}
We first demonstrate averaged model performance on all clients, which shows that a global model learned with FedCS will achieve comparable performance to other personalized FL methods that require model adaptation. Then, we look inside the group-wised metrics to evaluate a model's performance on different distributions. The global model learned with FedCS is more robust to different distributions. It can be directly deployed on test clients without extra adaptation. Several FL strategies are compared as baselines: 
1) FedAvg+FT~\cite{cheng2021fine}; 2) FedAvg+BN; 3) FedBN~\cite{li2021fedbn}. 
4) FedRep~\cite{collins2021exploiting}; 5) PerFL~\cite{pillutla2022federated}; 6) In addition, we demonstrate the performance of models those trained on each client locally (Local Only).

\subsection{Label-shift Settings}
\subsubsection{Overall Performance}
For label-shift settings, the weighted AUC score and the weighted F1 score are applied to evaluate model performance on data with unbalanced label distributions.
Table.\ref{table-averaged-mnist} demonstrates models' performance on the MNIST dataset. We can find that a model with FedCS layers achieves the best performance under the label-shift setting. It outperforms those locally fine-tuned global models (FedAvg+FT) and models with client-specific parameters (FedBN). FedRep has the worst performance, which may result from the lack of training data and the unbalanced label distribution of each client.

\begin{table}[t]
\centering
\caption{The averaged performance on the MNIST dataset. The standard deviation of the metric between clients is reported in parentheses. w. is the abbreviation of 'weighted', and the $\uparrow$ denotes that the higher the metric is, the better performance a model achieved, and the best performance is highlighted.}
\label{table-averaged-mnist}
\begin{tabular}{l|c|c}
    ~            & w. AUC ($10^{-2}$) $\uparrow$     & w. F1 ($10^{-2}$) $\uparrow$ \\
    \hline
    Local Only          & 85.18(4.62)    &47.80(10.12)\\
    FedAvg+FT           & 96.97(2.54)    &80.81(10.14) \\
    FedAvg+BN           & 99.69(0.14)    &93.38(1.76) \\
    FedBN               & 99.32(0.75)    &83.85(18.82)\\
    FedRep              & 75.72(15.56)   &38.53(20.52)\\
    PerFL               & 99.48(0.20)    &91.40(1.95)\\ 
    \hline
    FedCS-FC1(ours)     & \bf{99.72(0.15)}    &\bf{93.72}(1.71)\\
    FedCS-FC2(ours)     & 99.71(0.16)    &93.43(1.78)\\
    FedCS-FC3(ours)     & 99.72(0.26)    &93.64(1.56)\\
\end{tabular}

\end{table}

Table.\ref{table-averaged-cifar10} demonstrates models' performance on the CIFAR-10 dataset. FedCS has the best performance under this setting. Other models are less effective than FedCS, and their performances vary significantly among clients (higher standard deviations). We will show that the gap results from the generalization error on test clients, and there is no such problem for the proposed FedCS.

\begin{table}[t]
\centering
\caption{The averaged performance on the CIFAR-10 dataset. The standard deviation of the metric between clients is reported in parentheses. w. is the abbreviation of 'weighted', and the $\uparrow$ denotes that the higher the metric is, the better performance a model achieved, and the best performance is highlighted.}
\label{table-averaged-cifar10}
\begin{tabular}{l|c|c}
    ~            & w. AUC ($10^{-2}$)$\uparrow$     & w. F1 ($10^{-2}$)$\uparrow$ \\
    \hline
    Local Only          & 70.93(9.23)    &34.86(10.71)\\
    FedAvg+FT           & 86.89(4.65)    &51.75(12.62) \\
    FedAvg+BN           & 91.57(4.30)    &57.87(14.95) \\
    FedBN               & 91.64(4.56)    &59.41(14.70)\\
    FedRep              & 70.70(9.80)    &33.01(13.88)\\
    PerFL               & 89.18(3.36)    &58.62(9.17)\\ 
    \hline
    FedCS-FC1(ours)     & \bf{93.72}(1.71)    &69.48(4.69)\\
    FedCS-FC2(ours)     & 93.33(2.33)    &69.06(5.97)\\
    FedCS-FC3(ours)     & 93.49(2.09)    &\bf{69.69(4.96)}\\
\end{tabular}

\end{table}

\begin{table}[t]
\centering
\caption{The averaged accuracy on the Digit-5 dataset. The standard deviation of the metric between clients is reported in parentheses. w. is the abbreviation of 'weighted', and the $\uparrow$ denotes that the higher the metric is, the better performance a model achieved, and the best performance is highlighted.}
\label{table-averaged-digit5}
\begin{tabular}{l|c|c}
    ~            & w. AUC ($10^{-2}$)$\uparrow$     & w. F1 ($10^{-2}$)$\uparrow$ \\
    \hline
    Local Only          & 84.03(9.75)    &49.73(20.38)\\
    FedAvg+FT           & 96.11(1.98)    &75.64(4.94) \\
    FedAvg+BN           & 97.82(3.11)    &83.40(12.96) \\
    FedBN               & 95.71(3.79)    &74.07(10.90)\\
    FedRep              & 82.94(11.54)   &50.14(20.88)\\
    PerFL               & 96.10(2.47)    &74.63(6.71)\\ 
    \hline
    FedCS-FC1(ours)     & \bf{98.74}(0.95)    &\bf{87.89}(4.76)\\
    FedCS-FC2(ours)     & 98.60(1.09)    &87.66(5.15)\\
    FedCS-FC3(ours)     & 98.57(1.09)    &87.66(5.19)\\
\end{tabular}

\end{table}



\subsubsection{Group-wised Performance}
Fig.\ref{fig-grouped-f1-score-cifar10} demonstrates averaged weighted F1-scores within each client group\footnote{Full version of results are shown in Appendix.}. We can find that the global model with FedCR is more robust among different clients, even if they are from test groups (groups 8-9). Fine-tuned models (FedAvg+FT) and models with personalized parameters (FedBN) have significant performance gaps when deployed on training clients (group 0-7) and test clients. They achieve higher F1 scores in training groups but could be less effective in test groups.

\subsection{Feature-shift Settings}
In this section, we demonstrate evaluations of feature-shifted data. According to Table.\ref{table-averaged-digit5}, we show that FedCS achieves the highest weighted AUC and weighted-F1 score. The averaged accuracy within each data domain is shown in the Appendix. 
It also validates our claims that FedCS performs more robustly on all domains while other methods degenerate significantly on test clients. 

\section{Conclusion}
This paper is the first to propose using a one-model-for-all strategy to implement personalized federated learning. We believe the one-model-for-all personalization can form a new topic to advance existing personalized federated learning research. It is foreseeing more discussion and exploration can be conducted in this new one-model-for-all personalized federated setting. 


\bibliographystyle{IEEEbib}
\bibliography{icme24_6_reference}

\newpage
\appendix

\section{Theory Backgrounds}
We introduce an inductive bias to align the hidden layers of a DNN so that it is able to learn client bias to achieve personalization without on-device fine-tuning. We assume data on the same client are influenced by the same client properties so that data from the similar clients will have similar representations. We formulate the representation alignment problem into an optimization described in Eq.\ref{def-rayleigh-quotient}.

Eq.\ref{def-rayleigh-quotient} is equivalent to the objective function of Linear Discriminant Analysis (LDA) whose solution is the enginvector of corresponding to the largest enginvalue of $\Sigma_W^{-1}\Sigma_B$. However, the computation cost of the decomposing $\Sigma_W^{-1}\Sigma_B$ would be high, and aggregating local representations to a server is infeasible in FL. Therefore, we divide the matrix decomposition process into a set of clients' local updating steps and integrate it into standard FL framework.

\subsection{Client Supervised Optimization}
Let $\Sigma_G$ denote the correlation matrix of latent representations on all clients, there is $\Sigma_G=\Sigma_W+\Sigma_B$. Then the learning problem can be formulate as solving the following eigenvalue problem
\begin{equation}
    \begin{aligned}
    \Sigma_{W}^{-1}\Sigma_G P^{*}=P^{*}\Lambda
    \end{aligned}
    \label{eq-lda}
\end{equation}
where $\Lambda$ is the diagonal eigenvalue matrix of $\Sigma_W^{-1}\Sigma_G$. \cite{ghassabeh2015fast} shows solving Eq.\ref{eq-lda} can be simplified into solving the following symmetric eigenvalue problem:
\begin{equation}
    \begin{aligned}
        \Sigma_{W}^{-1/2}\Sigma_{G}\Sigma_{W}^{-1/2}\Phi=\Phi\Lambda
    \end{aligned}
    \label{eq-sym}
\end{equation}
where $\Phi$ denotes eignevectors of $\Sigma_{W}^{-1/2}\Sigma_{G}\Sigma_{W}^{-1/2}$, and there is $P^{*}=\Sigma_{W}^{-1/2}\Phi$. To find the optimum $P^{*}$ is to find $\Phi$ and $\Sigma_{W}^{-1/2}$.

\cite{572105} introduced an incremental algorithm to optimize $\Phi$ and $\Sigma_{W}^{-1/2}$ through which we can distribute the optimizing steps to clients. Concretely, it proves that, 1)
\begin{equation}
    \begin{aligned}
        \Phi_{k+1} = \Phi_{k} + \lambda(\mathbf{z}_k\mathbf{z}_k^{T}\Phi_k-\Phi_k\tau[\Phi^{T}_k\mathbf{z}_k\mathbf{z}_k^T\Phi_k])
    \end{aligned}
\end{equation}
will converge to the enginvector matrix $\Phi$ when there are sufficient instance $\mathbf{z}_k$ sampled from the data distribution; 2). let $\mathbf{S}$ denotes $\Sigma_{W}^{-1/2}$, then
\begin{equation}
    \begin{aligned}
        S_{k+1} = S_{k} + \eta * (I - S_{k}\Sigma_WS_{k})
    \end{aligned}
    \label{eq-S}
\end{equation}
where $S_{k+1}$ will converge to the inverted square root of $\Sigma_{W}$ when 1) $S_{0}$ is initialized as a symmetric positive define matrix, and 2) there are sufficient instance $\mathbf{z}_k$ sampled from the data distribution. 

We apply the above methods in Eq.\ref{eq-phi} and Eq.\ref{eq-sigma} to update $P$ on each clients individually and aggregate local updates to align axis on different clients.

\subsection{Discussion on Privacy Protection}
According to the section above, clients requires to share local correlations to update the matrix $\Sigma_W$. However, $\Sigma_W$ is a global statistic where a client's local bias would be eliminated, privacy-protection methods like differential privacy are feasible to avoid privacy leakage.

\section{Experiments}
\subsection{Heterogeneity Settings}
Three benchmark datasets are applied to evaluate FedCR's performance.
\subsubsection{Label-shift}
\begin{figure}[ht]
\begin{center}
\centerline{\includegraphics[width=\columnwidth]{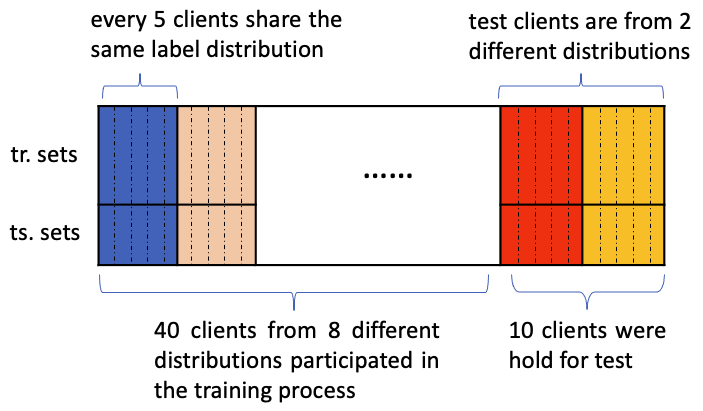}}
\caption{client settings for label shift experiments} 
\label{fig-client-settings}
\end{center}
\end{figure}

\begin{figure*}[ht]
\vskip 0.2in
\begin{center}
\centerline{\includegraphics[width=2.0\columnwidth,height=10cm]{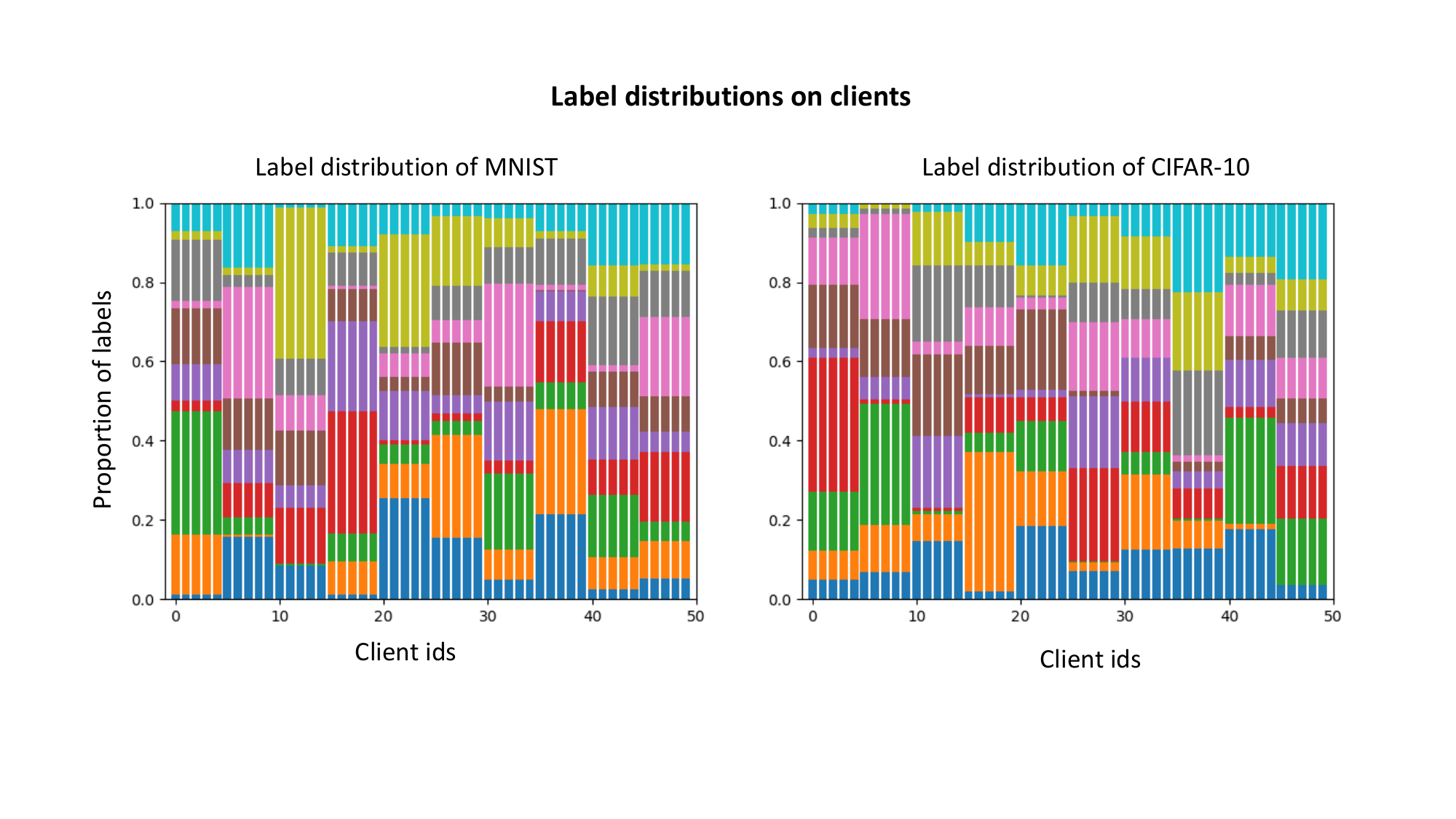}}
\caption{Proportion of instances of different classes. Different classes are marked with different colors. Horizontal axix denotes client ids, and the vertical axis denotes the proportion of classes} 
\label{fig-label-dist}
\end{center}
\vskip -0.2in
\end{figure*}

We allocate instances of each label individually according to a posterior of the Dirichlet distribution\cite{hsu2019measuring}, which divides clients into ten groups with different label distributions. Eight groups of clients will participate in the collaborative training process, and the rest will be held for testing. An example of the client setting is in Fig.\ref{fig-client-settings}. Distributions of the number of instances of each class are show in Fig.\ref{fig-label-dist}

\subsubsection{Feature-shift}
\begin{figure}[ht]
\begin{center}
\centerline{\includegraphics[width=\columnwidth]{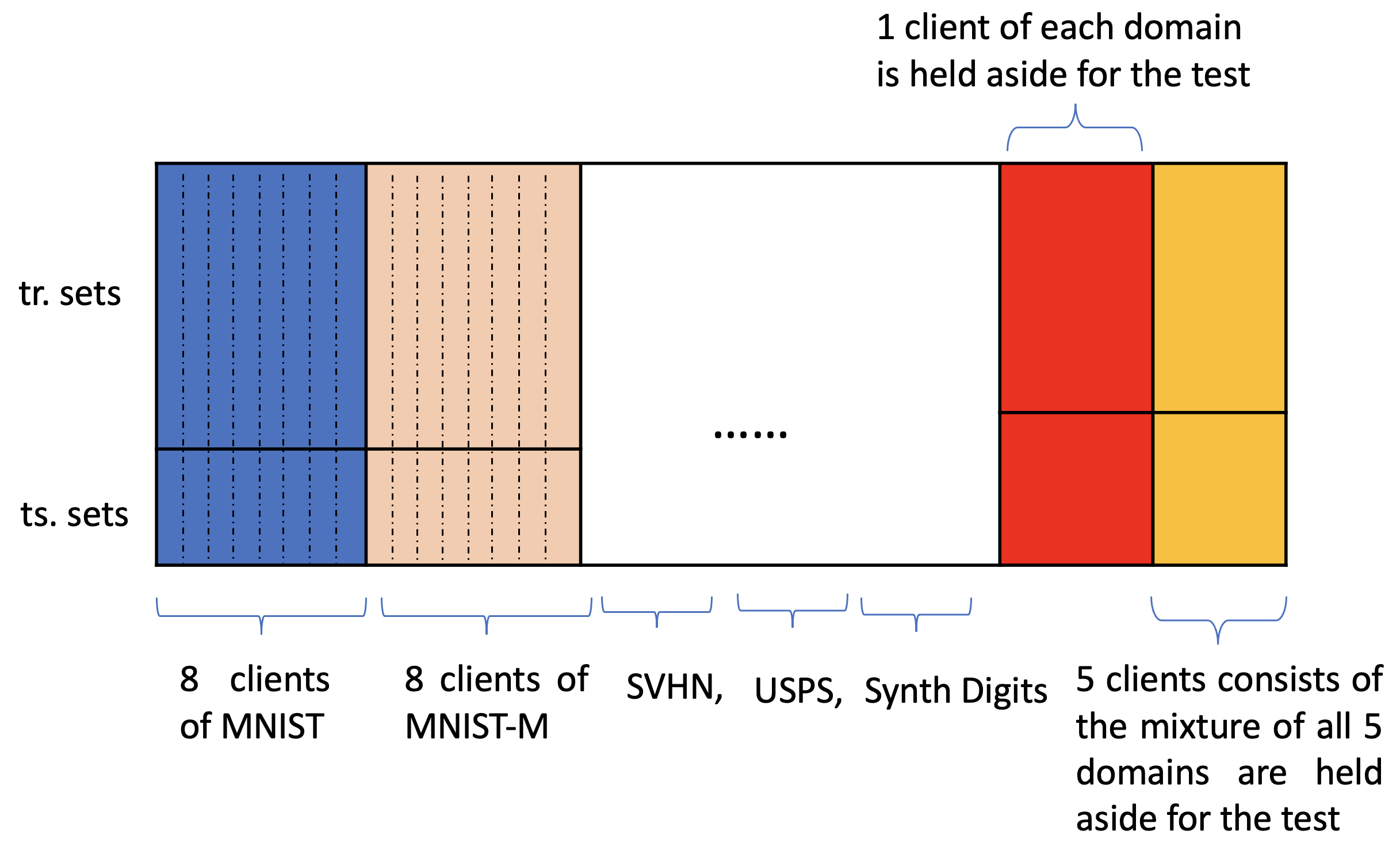}}
\caption{client settings for feature shift experiments} 
\label{fig-feature-dist}
\end{center}
\end{figure}

We experiment on the Digit-5 dataset to evaluate FedCRR's performance on feature-shift data. The Digit-5 consists of digits from five different domains (MNIST, MNIST-M, SVHN, USPS and Synth Digits). We assign instances of each domain to nine clients, where eight clients will train the global model and one for the test. In addition, we randomly draw instances from all domains to compose five mixed datasets for the rest clients for the test. The distributions on clients are shown in Fig.\ref{fig-feature-dist}

\subsection{Metrics}
We apply three different metrics to evaluate a model's performance on each clients. They are: accuracy, weighted F1 score and weighted AUC. We apply implementations in scikit-learn\footnote{https://scikit-learn.org/stable/index.html} in our experiments and utilize Laplacian smooth for clients where some labels are missing.

\subsection{Model Architecture}
Please refers attached codes for details of our experiments.

\subsection{Supplementary Results}
In addition to the results in Sec.\ref{Experiments}, more results are demonstrated as below. We can find our methods achieve better robustness and performance, especially on unseen/test clients.

\begin{figure}[t]
    \begin{center}
        \epsfig{figure=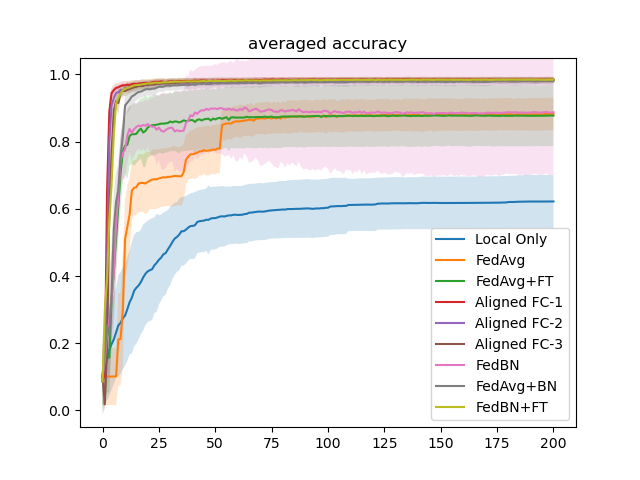,width=8.5cm}
    \end{center}
  \caption{Accuracy on the MNIST dataset. y-axis denotes the averaged accuracy on all clients and x-axis denotes the communication round. Shades denote the standard deviation of accuracy among clients. Our FedCS (Aligned FC-1 to Aligned FC3) achieves the best and the most robust performance.}
  \label{fig:res_mnist}
\end{figure}

\begin{figure}[t]
    \begin{center}
        \epsfig{figure=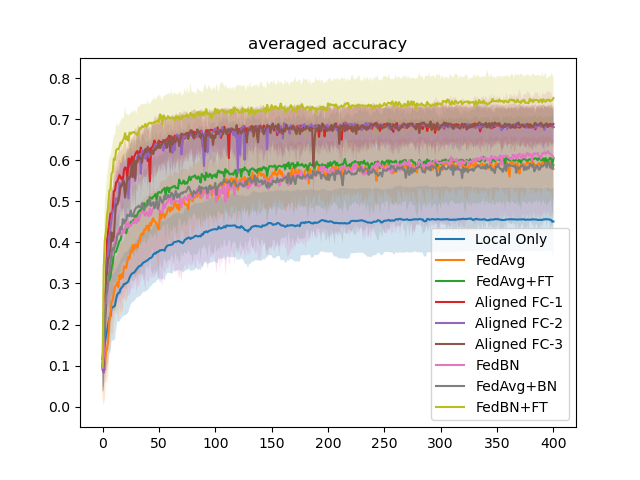,width=8.5cm}
    \end{center}
    \caption{Accuracy on the CIFAR-10 dataset. y-axis denotes the averaged accuracy on all clients and x-axis denotes the communication round. Shades denote the standard deviation of accuracy among clients. Our FedCS (Aligned FC-1 to Aligned FC3) achieves the best and the most robust performance.}
  \label{fig:res_cifar10}
\end{figure}

\begin{figure}[t]
    \begin{center}
        \epsfig{figure=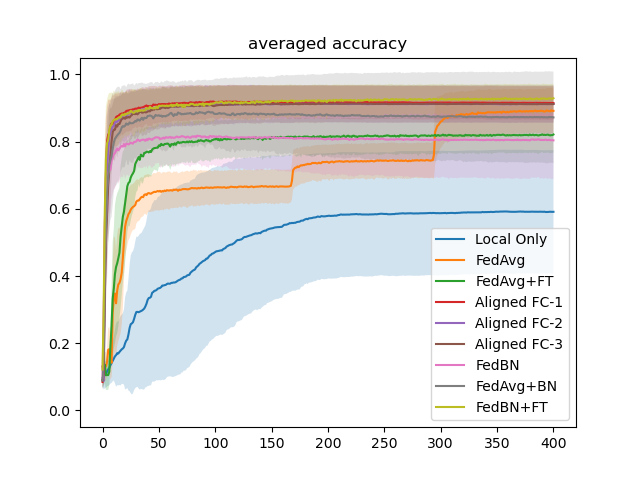,width=8.5cm}
    \end{center}
    \caption{Accuracy on the Digit-5 dataset. y-axis denotes the averaged accuracy on all clients and x-axis denotes the communication round. Shades denote the standard deviation of accuracy among clients. Our FedCS (Aligned FC-1 to Aligned FC3) achieves the best and the most robust performance.}
  \label{fig:res_digit5}
\end{figure}

\begin{figure*}[ht]
    \begin{center}
    \includegraphics[width=2.0\columnwidth,]{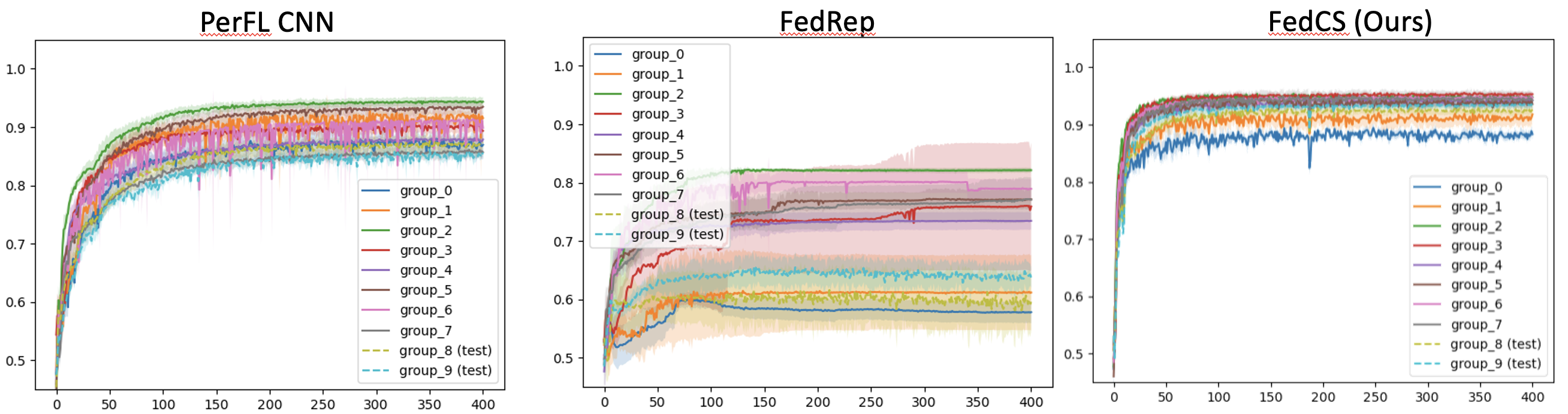}
    \caption{Part of experiments on CIFAR-10. The horizontal axis denotes communication rounds, and the vertical axis denotes accuracy. We demonstrate the averaged accuracy within each client group and marked them with different colours. We can find the performance on different distributions (client groups) vary significantly through FedAvg+FT and FedBN. FedCS has the most robust performance even on unseen clients (group 8 and 9)}
    \end{center}
  \label{fig:res_grouped_cifar10}
\end{figure*}

\begin{figure*}[ht]
    \begin{center}
    \includegraphics[width=2.0\columnwidth,]{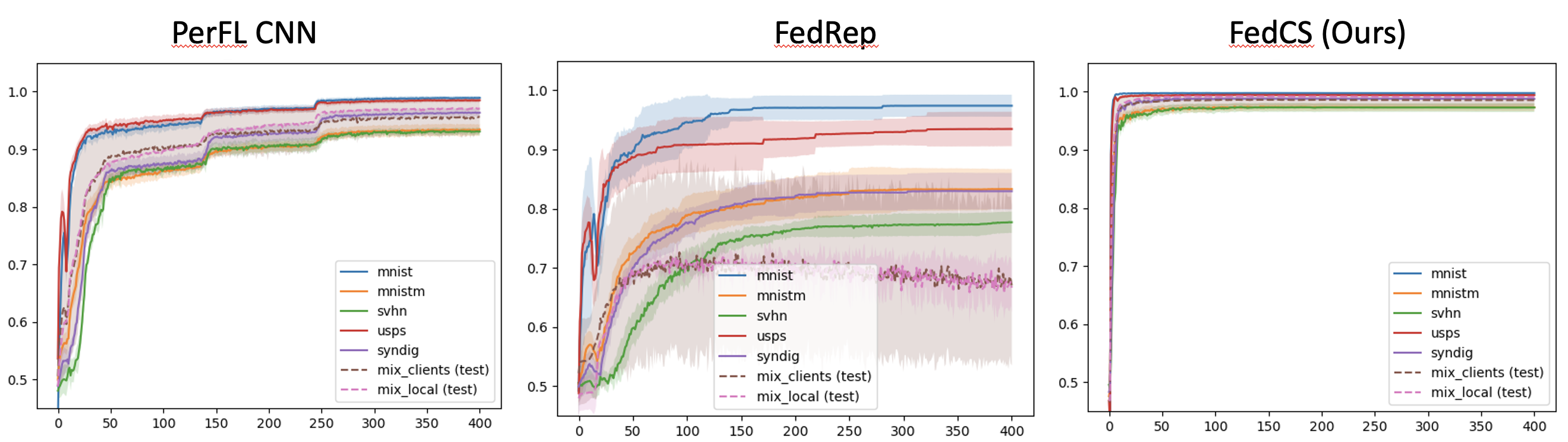}
    \caption{Part of experiments on Digit5. The horizontal axis denotes communication rounds, and the vertical axis denotes accuracy. We demonstrate the averaged accuracy within each client group and marked them with different colours. We can find the performance on different distributions (client groups) vary significantly through FedAvg+FT and FedBN. FedCS has the most robust performance even on unseen clients (group 8 and 9)}
    \end{center}
  \label{fig:res_grouped_digit5}
\end{figure*}

\end{document}